\documentclass[sigconf]{acmart}

\AtBeginDocument{%
  \providecommand\BibTeX{{%
    \normalfont B\kern-0.5em{\scshape i\kern-0.25em b}\kern-0.8em\TeX}}}

\copyrightyear{2025}
\acmYear{2025}
\setcopyright{acmlicensed}\acmConference[MM '25] {Proceedings of the 33rd ACM International Conference on Multimedia}{October 27--31, 2025}{Dublin, Ireland.}
\acmBooktitle{Proceedings of the 33rd ACM International Conference on Multimedia (MM '25), October 27--31, 2025, Dublin, Ireland}
\acmDOI{10.1145/3746027.3755714}
\acmISBN{979-8-4007-2035-2/2025/10}

\usepackage{balance}
\usepackage{algorithm}
\usepackage{algpseudocode}
\usepackage{subfig}
\usepackage{hyperref}
\usepackage{multirow}
\usepackage{booktabs}
\usepackage{amsthm,amsmath}
\usepackage{mathrsfs}
\usepackage{graphicx}
\usepackage{color,xcolor}
\usepackage{soul}
\usepackage{enumitem}
\usepackage{algpseudocode}
\usepackage{bbding}
\usepackage{adjustbox}
\definecolor{myorange}{RGB}{197, 90, 17}
\definecolor{mygreen}{RGB}{84, 130, 53}
\definecolor{myred}{rgb}{0.69, 0.25, 0.21}
\definecolor{redbackground}{rgb}{0.99, 0.90, 0.90}
\definecolor{greenbackground}{RGB}{226, 240, 217}
\definecolor{orangebackground}{RGB}{251, 229, 214}
\definecolor{codegreen}{RGB}{0,176,80}
\definecolor{codegray}{rgb}{0.5,0.5,0.5}

\usepackage{newfloat}
\usepackage{listings}
\usepackage{graphicx}

\begin{document}

\title{ChartM\texorpdfstring{$^3$}{3}: Benchmarking Chart Editing with Multimodal~Instructions}


\settopmatter{authorsperrow=4}

\author{Donglu Yang}
\email{2022201826@ruc.edu.cn}
\orcid{0009-0003-7763-0094}
\affiliation{
  \institution{RUC}
  \city{Beijing}
  \country{China}
}

\author{Liang Zhang}
\email{zhangliang04@hotmail.com}
\orcid{0000-0002-6187-3628}
\affiliation{
  \institution{independent researcher}
  \city{Beijing}
  \country{China}
}

\author{Zihao Yue}
\email{yzihao@ruc.edu.cn}
\orcid{0009-0003-7763-0094}
\affiliation{
  \institution{RUC}
 \city{Beijing}
  \country{China}
}
\author{Liangyu Chen}
\email{liangyuchen@ruc.edu.cn}
\orcid{0000-0003-2099-4025}
\affiliation{
  \institution{RUC}
  \city{Beijing}
  \country{China}
}
\author{Yichen Xu}
\email{xu_yichen@ruc.edu.cn}
\orcid{0009-0005-8113-6464}
\affiliation{
  \institution{RUC}
 \city{Beijing}
  \country{China}
}
\author{Wenxuan Wang\textsuperscript{*}}
\email{jwxwang@gmail.com}
\orcid{0000-0002-9803-8204}
\affiliation{
  \institution{RUC}
  \city{Beijing}
  \country{China}
}
\author{Qin Jin}
\email{qjin@ruc.edu.cn}
\orcid{0000-0001-6486-6020}
\affiliation{
  \institution{RUC}
  \city{Beijing}
  \country{China}
}
\authornote{Qin Jin and Wenxuan Wang are corresponding authors.}


\begin{abstract}
Charts are a fundamental visualization format widely used in data analysis across research and industry. While enabling users to edit charts based on high-level intentions is of great practical value, existing methods primarily rely on natural language instructions, which are often too ambiguous to support fine-grained editing. In this work, we introduce a novel paradigm for multimodal chart editing, where user intent is expressed through a combination of natural language and visual indicators that explicitly highlight the elements to be modified.
To support this paradigm, we present Chart$\text{M}^3$, a new benchmark for Multimodal chart editing with Multi-level complexity and Multi-perspective evaluation. Chart$\text{M}^3$ contains 1,000 samples spanning four levels of editing difficulty. Each sample includes triplets in the form of (chart, code, multimodal instructions). To comprehensively evaluate chart editing models, Chart$\text{M}^3$ provides metrics that assess both visual appearance and code correctness.
Our benchmark reveals significant limitations in current multimodal large language models (MLLMs), including GPT-4o, particularly in their ability to interpret and act on visual indicators. To address this, we construct Chart$\text{M}^3$-Train, a large-scale training set with 24,000 multimodal chart editing samples. Fine-tuning MLLMs on this dataset leads to substantial improvements, demonstrating the importance of multimodal supervision in building practical chart editing systems. Our datasets, codes, and evaluation tools are available at \url{https://github.com/MLrollIT/ChartM3}.

\end{abstract}





\maketitle
\begin{figure}[H] 
\centering \includegraphics[width=\linewidth]{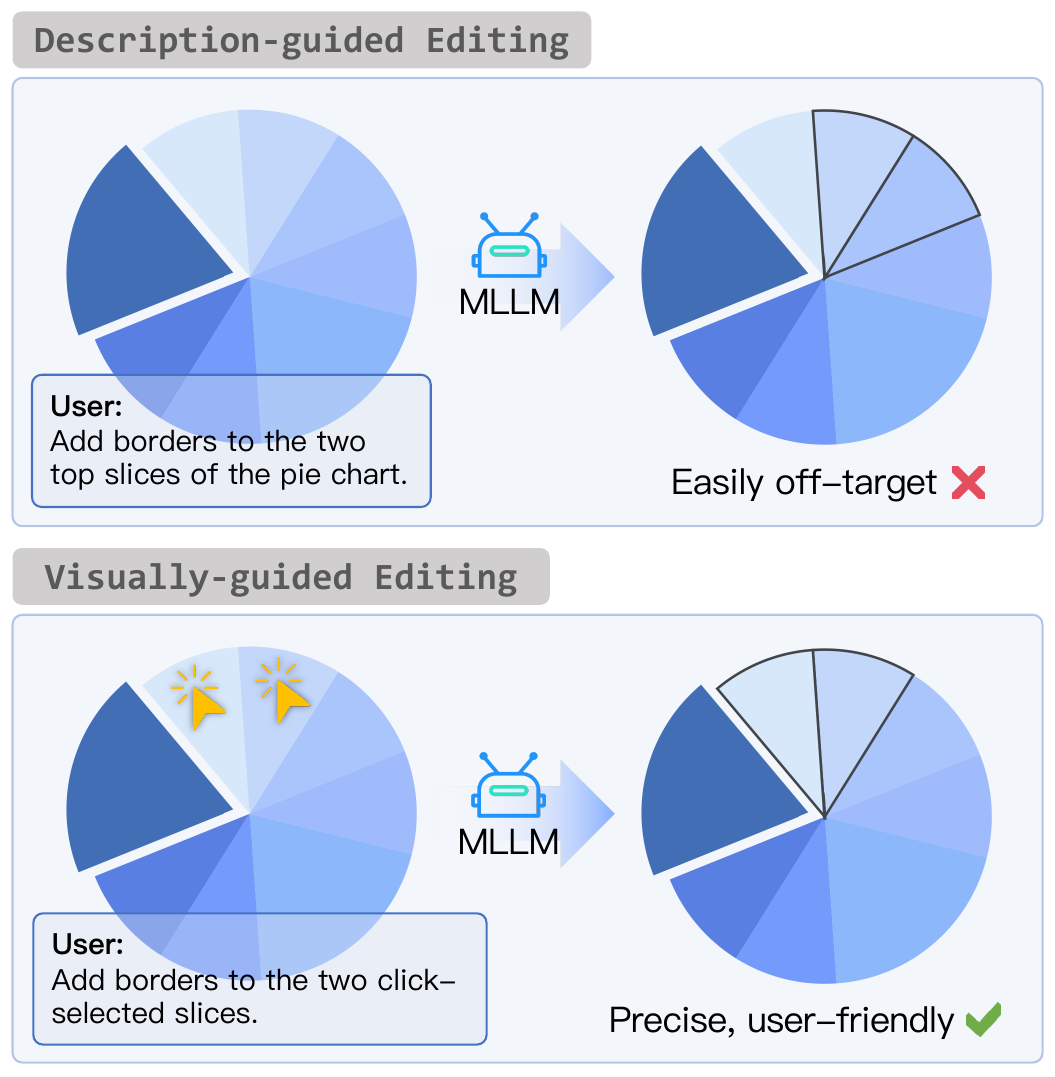} 
\caption{An example of text description-guided and visual indicator-guided chart editing. The model mistakenly identified the target slice with description-guided chart editing, and it could correctly identify with visual indicators.} 
\label{fig:intro}
\end{figure}
\section{Introduction}

Charts serve as a powerful medium for communicating structured information across domains ranging from scientific publications and business reports to educational materials and user presentation dashboards. As chart generation becomes increasingly automated, there is a growing demand for tools that can support interactive chart editing—enabling users to revise visualizations with minimal effort and maximal precision.
Chart editing~\cite{Yang2025ChartMimic} has therefore received increasing attention.

Unlike natural images, visual elements in charts, such as bars, lines, and legends, are governed by a strict internal logic and must maintain semantic consistency and data integrity during modifications. Thus, traditional image editing approaches that modify the image at the pixel level~\cite{9879284,10204579,hertz2022prompt} have been proved to be inadequate for chart editing, since they lack understanding of the structured nature of chart images. Recent studies in chart editing~\cite{Yan2024ChartReformer} research predominantly relies on multi‐stage workflows that require specialized tools to extract chart elements, thereby precluding fully end‐to‐end editing capabilities; moreover, the spectrum of supported chart types and available modification options remains severely constrained.

In practice, users often express their editing intentions through natural language (e.g., “change the color of the third bar to red”). However, such instructions are inherently ambiguous and often fail to specify precise visual targets, especially for fine-grained edits like adjusting a specific segment in a stacked bar chart or switching positions between two pie slices. As shown in the upper part of Figure~\ref{fig:intro}, due to the ambiguity of natural language, the description-guided instruction causes the model to produce a wrong modification in visual elements.
Such ambiguity poses a significant challenge for Multimodal Large Language Models (MLLMs), which must infer and localize the intended visual changes and translate them into accurate code edits.

To overcome this limitation, we propose a multimodal editing framework that combines both natural language instructions and visual indicators to more precisely convey user intent. 
As shown in the lower part of Figure~\ref{fig:intro}, by identifying the visual elements with visual indicators\footnote{Note that in practice, we automatically draw a bounding box around the user's click position as the visual indicators.}, the model could correctly find the elements to be modified and perform corresponding modifications. 
This hybrid approach bridges the gap between abstract textual descriptions and concrete visual cues, enabling more accurate alignment between user goals and programmatic edits.

To support the systematic study of the proposed multimodal chart editing problem, we introduce Chart$\text{M}^3$, a novel benchmark for \textbf{M}ultimodal chart editing with \textbf{M}ulti-level complexity and \textbf{M}ulti-perspective evaluation. Chart$\text{M}^3$ consists of 1,000 carefully curated quadruplets in the form of (chart, code, multimodal prompt, instruction), covering a wide range of chart types and editing tasks. We define two complementary editing paradigms within this benchmark to evaluate MLLMs' ability to bridge multimodal chart understanding and code modification. Specifically, the two distinct editing paradigms are: 1) \textbf{Textual description-guided editing}, where models are given natural language descriptions to identify to-be-modified elements within the chart before implementing appropriate code modifications. (e.g., "the third bar from the left"). This tests a model's foundational ability to understand semantic descriptions and map them to visual elements. 2) \textbf{visual indicator-guided editing}, where models receive visual indicators in the form of bounding boxes highlighting the to-be-modified target regions. While this setting reduces descriptive ambiguity, it presents a more demanding challenge for models, demanding advanced visual-to-code mapping capabilities as they must recognize visual indicators and associate visual regions with code-level constructs.

Chart$\text{M}^3$ also features detailed annotations and metadata, capturing element-level correspondences between visual components and code entities. We further design multidimensional evaluation metrics that assess models along several axes: target localization accuracy, code correctness, and visualization integrity after code modification.

Using Chart$\text{M}^3$, we evaluate a broad set of MLLMs~\cite{openai2023gpt4v, wang2024qwen2vl, Lee2025TrimmedLlama, Zhao2025ChartCoder,liu2024llavanext,internvl2,internvl2_5,lu2024deepseekvl} and reveal notable limitations in their current capabilities, reflecting significant challenges across both complexity levels in our proposed benchmark. In the textual description-guided editing paradigm, models frequently misinterpret textual descriptions, identifying incorrect elements or generating modifications to unintended parts. More surprisingly, even with explicit visual indicator-guided visual indicators, models struggle to correctly associate these indicated regions to their corresponding code representations, highlighting a major gap in multimodal-to-code translation. 

To advance this area, we curate a large-scale training dataset of 24,000 samples that simulate a diverse set of multimodal chart-editing scenarios. Fine-tuning existing models on this dataset yields substantial performance gains across both textual and visual guidance modes, setting new state-of-the-art baselines for this task.

In summary, our key contributions are as follows: 
\begin{itemize}[leftmargin=*]
\item \textbf We introduce a new paradigm for chart editing that combines natural language instructions with explicit visual indicators. This approach addresses the limitations of purely text-based editing methods by allowing users to precisely highlight the elements they wish to modify, significantly reducing ambiguity in expressing fine-grained editing intentions.
\item \textbf We present Chart$\text{M}^3$, a comprehensive benchmark for evaluating multimodal chart editing capabilities. Chart$\text{M}^3$ features 1,000 diverse samples in four levels of editing complexity. Moreover, Chart$\text{M}^3$ includes a multi-dimensional evaluation framework that assesses both visual appearance and code correctness.
\item \textbf Through extensive experiments on Chart$\text{M}^3$, we demonstrate that existing multimodal large language models struggle to align structured data across modalities, particularly when handling visual indicators including bounding boxes, revealing fundamental limitations in their ability to translate user's visual intension into corresponding code.
\end{itemize}

\begin{figure*}[t] 
\centering 
\includegraphics[width=\textwidth]{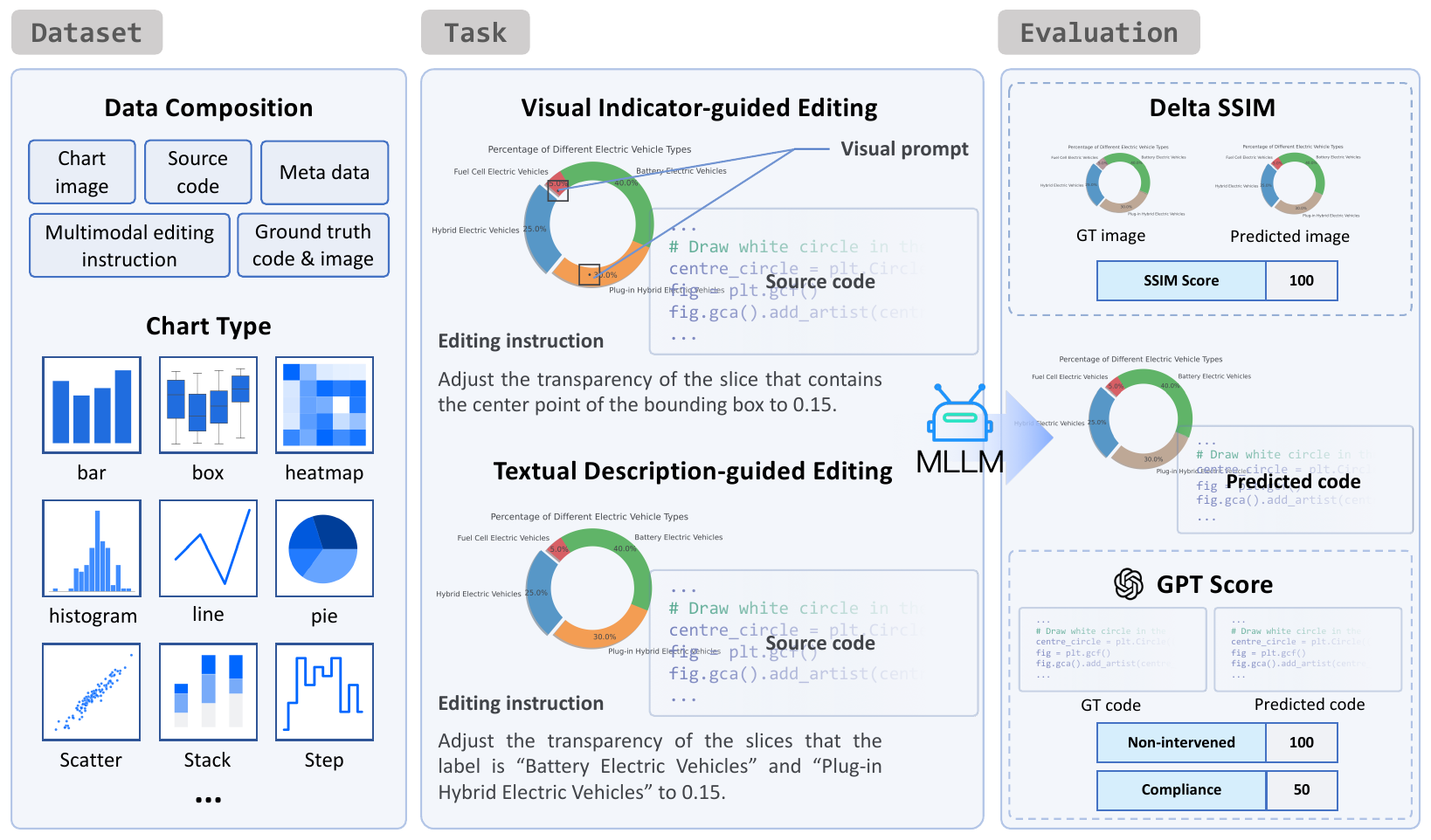} 
\caption{The overview of Chart$\text{M}^3$ (Multimodal, Multilevel Complexity, Multidimensional Evaluation). Our benchmark features 1,000 human-curated (chart, code, multimodal instruction) triplets that evaluate MLLMs' proficiency in chart modification tasks across multiple complexity levels, incorporating both high-level textual descriptions and low-level visual indicator-guided editing instructions.} 
\label{fig:example} 
\end{figure*}

\section{Related Works}

\noindent\textbf{{Chart-related Research.}}
As a specific image specific type of image for data visualization, charts have attracted increasing research attention. Existing works explore both chart 
understanding and generation tasks. For example, chart understanding tasks such as chart question answering~\cite{Chen2025ChartHQA,Masry2022ChartQA,Methani2019PlotQA,Kantharaj2022OpenCQA,Vogel2025RefChartQA}, chart-to-table, and chart captioning mainly focus on semantic understanding of chart content. Chart generation tasks such as text-to-chart~\cite{PesaranZadeh2024Text2Chart31} and chart-to-code~\cite{Zhao2025ChartCoder} generation focus on automatic chart creation. Recent works~\cite{Yan2024ChartReformer,goswami2025plotedit,Yang2025ChartMimic} propose the chart editing task that enables modification of specific chart elements based on user intention. It enables interactive chart creation with users. 
Though progress has been made, these works represent user intention primarily with natural language instructions, which can hardly identify fine-grained manipulation of specific chart elements due to ambiguity. Apart from natural language instructions, our proposed Chart$\text{M}^3$ provides visual indicators to precisely identify the elements to be edited. It also offers a fresh perspective to evaluate a model's understanding of both chart images and their corresponding code. 

\noindent\textbf{Image Editing.}
Image editing evolves from pixel-level manipulation tasks, such as inpainting~\cite{Pathak2016ContextEncoders} and style transfer~\cite{Ghiasi2017Stylization}, to more advanced text-based~\cite{Kawar2022Imagic,sheynin2023emu} and drag-based~\cite{pan2023draggan,shi2023dragdiffusion} editing based on diffusion models~\cite{SohlDickstein2015DeepUnsupervised} and generative adversarial networks (GAN)~\cite{Goodfellow2014GANs}. These approaches primarily focus on making visual modifications based on textual prompts or user-provided drag instructions. They excel in editing natural images or artistic images, producing creative and consistent results. 
In contrast to pixel-level editing, chart editing requires the model to strictly preserve visual elements that are not specified to be edited, and thus requires editing based on its underlying code. It thus presents an additional challenge to the model to translate the visual intention into appropriate code modification. 

\noindent\textbf{Multimodal Large Language Models.} 
Research in Multimodal large language models (MLLMs) achieved significant progress in recent years. For example, close-sourced MLLMs such as GPT-4V~\cite{openai2023gpt4v} is capable of completing complex real world vision and language tasks. Open-source MLLMs such as LLaVA~\cite{liu2023visual,Xu2024LLaVA-UHD}, Qwen-VL~\cite{wang2024qwen2vl}, and DeepSeek-VL~\cite{lu2024deepseekvl} also demonstrate good performance compared to previous method in many standard multimodal understanding benchmarks. However, these models frequently underperform close-sourced models in real-world scenarios. Thus, it is necessary to construct new evaluation frameworks that reflect practical applications. Our ChartM$^3$ benchmark resolves this demand by evaluating MLLMs' capability with chart editing based on visual indicators and textual instructions. It assesses the capacity to perform visual language understanding and generate functional code, and provides a more user-centric evaluation setting that closely aligns with real-world usage scenarios. 

\section{The ChartM³ Benchmark}%

In this section, we introduce our ChartM$^3$ benchmark including its task definition, data construction process, and our proposed evaluation metrics.
\begin{table*}[h]
  \caption{Data statistics across chart types, showing dataset size, code characteristics, and instruction properties for both task formats. We report code length (average number of characters), code diversity (measured by 3-gram uniqueness), instruction length (average word count), and instruction similarity (cosine similarity between instructions).}
  \label{tab:data_analysis}
  \centering
  \begin{tabular}{lccccccccccc}
    \toprule
    Statistics & Bar & Box & Heatmap & Histogram & Line & Pie & Scatter & Stack & Step & Violin & Avg\\
    \midrule
    Total Size & 100 & 125 & 50 & 100 & 100 & 125 & 50 & 125 & 100 & 125 & 100 \\
    Code Length & 1135 & 1241 & 1220 & 666 & 1385 & 1040 & 983 & 742 & 714 & 908 & 1003 \\
    Code Diversity & 0.56 & 0.29 & 0.46 & 0.65 & 0.45 & 0.39 & 0.57 & 0.69 & 0.70 & 0.62 & 0.54 \\
    \midrule
    Task1 Instruction Length & 132.2 & 156.5 & 294.7 & 166.6 & 116.0 & 154.8 & 164.4 & 148.3 & 126.1 & 153.0 & 161.2 \\
    Task1 Instruction Similarity & 0.498 & 0.454 & 0.621 & 0.409 & 0.470 & 0.441 & 0.525 & 0.508 & 0.420 & 0.492 & 0.484 \\
    \midrule
    Task2 Instruction Length & 72.4 & 104.9 & 91.0 & 101.5 & 73.8 & 86.01 & 85.5 & 97.7 & 87.7 & 88.4 & 88.9 \\
    Task2 Instruction Similarity &0.287 & 0.316 & 0.493 & 0.243 & 0.239 & 0.312 & 0.451 & 0.356 & 0.301 & 0.318&0.332 \\
    
    \bottomrule
  \end{tabular}
\end{table*}

\subsection{Task Definition}

\noindent{\textbf{Chart Editing.}} Chart editing aims to modify the content of a chart image based on user intention. Traditional chart editing task conveys user intention by natural language instructions. Since charts are typically generated by a specific programming language, a chart is represented with both the chart image and it's underlying code. More formally, given the original chart image $I$, the corresponding chart code $c$, and the textual instruction $T$, chart editing task can be represented as follows:
\begin{align}
\hat{c} = \mathrm{M}(I,T,c),
\end{align}
where $\hat{c}$ is the underlying code of the editing result image $\hat I$, and $\mathrm{M}$ is the chart editing model.

\noindent\textbf{Multimodal Chart Editing.} In contrast to traditional chart editing, our proposed ChartM$^3$ benchmark utilizes multimodal guidance to identify the visual elements to be modified. Specifically, we consider two types of guidance to identify the user intention: the \textbf{textual-description guidance} and the \textbf{visual-indicator guidance}. Specifically, the textual description contains detailed natural language instruction that not only describes the desired modification but also provides textual references to identify the specific chart elements that should be modified. In contrast, the
visual indicator contains a group of bounding box annotations inside the original chart image to highlight the specific elements to be modified. Note that this bounding box can be automatically generated around the user click position to make this process more friendly. More formally, given a textual description guidance $T_\text{desc}$ or a visual indicator $vi$, the multimodal chart editing task designed in ChartM$^3$ can be represented by the following formula:

\begin{align}
\hat c = \begin{cases}
\mathrm{M}(I,c,T_\text{desc}) & \text{for Task 1: Textual Description-guided Editing,} \\
\mathrm{M}(I^{vi}, c, T) & \text{for Task 2: Visual Indicator-guided Editing,} \\
\end{cases}
\end{align}
where $I^{vi}$ is the original chart image with the visual indicator $vi$. $T$ is the basic task instruction that describes the desired chart modification without identifying the elements.

\noindent These well-structured levels enable a comprehensive evaluation of MLLMs' chart editing capabilities, assessing their ability to process both explicit visual annotations and textual element descriptions. Additionally, this dual-format approach mirrors real-world interactions where users either point to specific regions or describe them verbally, providing critical insights for developing more effective chart editing systems.

\noindent \textbf{Multi-level Complexity.}
To comprehensively and meticulously evaluate model performance, we further split the task samples into four progressive difficulty levels based on the number of modification targets and instructions. Specifically, we categorize the samples in ChartM$^3$ benchmark into: single-target single-instruction (SS), multi-target single-instruction (MS), single-target multi-instruction (SM), and multi-target multi-instruction (MM). This enables us to create a comprehensive evaluates the models' capabilities across different levels of chart understanding and code generation.


\subsection{Data Construction}
We developed a pipeline to create high-quality datasets for visual indicator-guided and text-guided chart editing tasks in ChartM$^3$. This includes sampling diverse chart types, generating modifications, creating dual-format instructions, and implementing quality controls. A more detailed description is provided in Section A of the supplementary material.

\noindent\textbf{Sampling and Code Generation.} We collected 10k chart code samples from ten categories. For each, we designed modification pools with Artist API calls and type-specific transformations. A two-stage procedure selected chart codes and applied 1-3 modifications, preserving integrity while creating diverse editing scenarios.

\noindent \textbf{Textual Instruction Generation.}
Instructions for Task 1 focused on modification operations, with target identification via bounding box annotations. Task 2 used position-based descriptions (e.g., "third bar from left") or semantic labels (e.g., "Revenue line").

\noindent \textbf{Automated Annotation and Quality Control.} An automated script generated bounding box annotations based on code modifications. GPT-4o validated generated images, filtering out those with clutter, ambiguous references, or imprecise annotations. This process created paired datasets for both task formats, balancing efficiency and rigor while minimizing manual effort.
\begin{table*}
  \caption{Evaluation results of MLLMs with the ChartM$^3$ benchmark.}
  \label{tab:model_performance}
  \centering
  \begin{tabular}{l|c|cccc|cccc}  
    \toprule
    \multirow{2}{*}{Model} & \multirow{2}{*}{Params}   & \multicolumn{4}{c}{\textbf{Textual Description}}& \multicolumn{4}{c}{\textbf{Visual Indicator}} \\ 
    \cmidrule(lr){3-6}  
    \cmidrule(lr){7-10}
    & & Execute Rate&iSSIM & Compliance&Non-inter & Execute Rate & iSSIM  & Compliance&Non-inter \\ \midrule
    \multicolumn{10}{c}{Zero-shot} \\ \midrule
    InternVL-2.5\cite{internvl2_5} & 0.938B& 17.7\% & 1.02&5.92&14.04 & 13.8\% & 0.86  & 3.93&10.23  \\
    ChartCoder\cite{Zhao2025ChartCoder}& 7.16B &71.2\%&12.98&28.14&66.20& 63.5\% &6.02&22.24&58.23 \\
    Deepseek-vl-chat\cite{lu2024deepseekvl} & 7.34B & 74.7\% &10.93&30.55&74.54  & 75.7\% & 6.91  & 24.57&68.43 \\
    InternVL-2\cite{internvl2} & 8.10B& 43.3\% & 10.53&23.60&39.74 & 39.9\% & 7.78&19.56&39.25   \\
    Qwen2-vl\cite{wang2024qwen2vl} & 8.29B & 72.5\% & 12.59&34.84&68.28& 60.7\% & 6.5 &23.58&55.66   \\
    Llama-3.2-vision\cite{Lee2025TrimmedLlama} & 10.70B& 69.7\% & 22.34&40.15&58.57 & 66.8\% & 11.18 &30.87&57.99  \\
    Llava-v1.6-vicuna\cite{liu2024llavanext} & 13.40B&   43.8\% & 7.50&23.52&39.53& 36.8\% & 4.23&17.17&30.85    \\
    
    GPT-4o\cite{openai2023gpt4v} & -& \textbf{90.3}\% &\textbf{56.54} &\textbf{76.80}&\textbf{87.44} & \textbf{87.2\%} & \textbf{38.08}  &\textbf{63.36}&\textbf{81.91}  \\
    \midrule
    \multicolumn{10}{c}{Supervised Fine-tuning} \\  
    \midrule
    Qwen2-vl\cite{wang2024qwen2vl} & 8.31B & 87.4\% & \textbf{63.23} &\textbf{73.06}
&85.35 & \textbf{86.5}\% & \textbf{57.88} & \textbf{71.04}&\textbf{84.45} \\
Llama-3.2-vision\cite{Lee2025TrimmedLlama} & 10.72B& \textbf{91.4\%} & 52.46&70.73&\textbf{90.4} & 83.4\% & 51.00 &61.76&82.13   \\
Llava-v1.6-vicuna\cite{liu2024llavanext} & 13.43B & 71.1\% & 35.41&53.00&70.02 & 71.1\%&35.37 & 52.01&70.00\\
    \bottomrule
  \end{tabular}
\end{table*}

\subsection{Data Statistics}
As shown in Table~\ref{tab:data_analysis}, our dataset comprises ten of the most common types encountered in data analysis. To account for the diversity of modification instructions, we deliberately controlled the number of data instances for each chart type, ranging from 50 to 125. Specifically, chart types such as heatmaps and scatter plots, which inherently support fewer modification operations, were assigned smaller counts, while chart types amenable to a broader range of modifications were allocated up to 125 instances. This design minimizes redundancy logic in modification, thereby enabling a more effective evaluation of the generalization capability. We also statistics the length, code diversity, and instruction similarity of the dataset. 
It reveals that our dataset is diversely distributed in terms of both code and instructions. Since we report 
This approach also facilitates subsequent analyses of individual type-specific results when utilizing this benchmark to evaluate model performance.

\subsection{Multi-perspective Evaluation Metrics}
ChartM$^3$ benchmark performs a multi-dimensional evaluation for the chart editing results including visual appearance perspective and code correctness perspective.

\noindent \textbf{$\Delta$SSIM.} We propose Delta Structural Similarity Index Measure ($\Delta$SSIM) to measure from the visual appearance perspective, by improving the SSIM~\cite{1284395} metric used in general image editing. SSIM~\cite{1284395} is widely regarded for its ability to capture subtle variations in structural, luminance, and contrast changes, making it an informative global metric for assessing the overall fidelity of generated outputs. However, standard SSIM alone does not account for the progressive improvement of a generated image ($\hat{I}_g$) over an initial input image ($I$) toward the reference image ($\hat I$). To remedy this, we propose an improvement to SSIM, referred to as the Delta SSIM ($\Delta$SSIM), which quantifies the relative enhancement achieved by the model during the transformation process.The $\Delta$SSIM score($s$), is specifically designed to measure the degree to which the generated image ($I_g$) approximates the reference image ($I_r$), normalized relative to the similarity of the initial image ($I_i$) to the reference image ($I_r$). It is computed as:
\begin{equation}
\Delta \mathrm{SSIM}(I_,\hat I,\hat I_g) =\mathrm{max}\left (0,\frac{S(\hat I_{g}, \hat I)-S(I, \hat I)}{1-S(I, \hat I)}*100\right ),
\end{equation}where \( S \) 
represents the SSIM value. The proposed $\Delta$SSIM has several advantages over the traditional SSIM metric: 1) It penalizes cases where the generated image fails to improve upon the initial image ($I_r$ is closer to the $I_i$ than $I_g$). 2) It provides a normalized measure of improvement that is invariant to the absolute values of the initial similarity $S(I_{i}, I_{r})$. Consequently, $\Delta$SSIM allows for a more nuanced evaluation of the model's ability to progressively enhance the input image toward the target, especially in cases where the initial similarity is low.

\noindent \textbf{GPT Score.}
Building on the successful application of large language models for automated evaluation, we utilize GPT-4~\cite{openai2023gpt4v} 
to evaluate model-generated chart modifications from the code correctness perspective. For each test case, we prompt GPT-4 with comprehensive evaluation context including: (1) the specific objects requiring modification (with both index and label information), (2) the modification instructions, (3) the model-generated code to be evaluated, and (4) the reference code serving as ground truth. Specifically, our GPT-4 evaluation framework assesses performance across two dual-dimensions:

\noindent \textbf{\emph{Directive Compliance Ratio.}} This dimension evaluates whether each target object has been correctly modified according to instructions. It considers two key aspects:(1) Whether all specified target objects in the generated code have been successfully modified with consistent results across all targets. (2) Whether modifications match the intended effect as demonstrated in the reference code, applying only the requested changes to the target objects.

\noindent \textbf{\emph{Non-intervened Robustness.}} This dimension assesses whether objects not marked for modification remained unchanged. We prompt GPT-4 to directly compare non-target elements in the generated code against the reference code. 

\noindent $\Delta$SSIM directly quantifies the visual-quality improvements introduced by the editing process by measuring the proportional gain in structural similarity between the generated output and the target, relative to the initial input's similarity. In contrast, the GPT Score operates at the semantic level-evaluating directive compliance and non-intervened robustness-to capture the model's reasoning and code-generation capabilities. Together, these two metrics offer complementary insights, with $\Delta$SSIM assessing perceptual fidelity and GPT Score probing the correctness and precision of the underlying modification logic.

\subsection{Supervised Fine-Tuning}
Existing chart editing frameworks~\cite{Chen2025ChartHQA,Yang2025ChartMimic,Masry2022ChartQA,Yan2024ChartReformer,Zhao2025ChartCoder,PesaranZadeh2024Text2Chart31,Vogel2025RefChartQA} rely on text-guided editing, limiting support for multimodal chart editing like in ChartM$^3$. To enhance its multimodal capabilities, we create the ChartM$^3$-Train dataset, which includes 24,000 samples using the same data construction pipeline from Section 3.2. This dataset contains an equal number of visual indicator-guided and textual description-guided samples.

The model processes instructions (basic or descriptive), images (annotated or unannotated), and initial code as input, generating reference code as output. We apply cross-entropy loss to minimize discrepancies between the generated and reference code, enabling the model to interpret visual annotations and textual references for accurate code modifications.

\section{Experiment}
\subsection{Evaluation Settings}
We benchmark eight leading multimodal large language models (both proprietary and open-source) on Chart$\text{M}^3$, including InternVL-2.5~\cite{internvl2_5}, ChartCoder~\cite{Zhao2025ChartCoder}, deepseek-vl-chat~\cite{lu2024deepseekvl}, Internvl-2~\cite{internvl2}, Qwen2-vl~\cite{wang2024qwen2vl}, Llama-3.2-vision~\cite{Lee2025TrimmedLlama}, llava-v1.6-vicuna~\cite{liu2024llavanext}, and GPT-4o~\cite{openai2023gpt4v}. We compare models on both zero-shot and fine-tuning settings, with the task prompt detailed in Appendix.

\subsection{Main Results} 

\noindent \textbf{Zero-shot Results.}
Table \ref{tab:model_performance} presents the performance of various models on the ChartM$^3$ benchmark test set. The proprietary GPT-4o achieves the highest execution rates, exceeding 87\% in both task settings, indicating strong reliability in generating valid code modifications. It also achieves high Compliance performance (76.80 in the textual description-guided setting and 63.36 in the visual indicator-guided setting) and Non-intervention Robustness (87.44 and 81.91, respectively), outperforming other models by a large margin and serving as a strong baseline for the chart editing task. Among open-source models, Llama-3.2-vision achieves the highest Compliance score and iSSIM score, indicating it best follows the editing instruction and yields the most visually similar outcomes. We also observe that models consistently perform better in the text-guided editing task than in the visually-guided one. This indicates that there is a notable gap for the models to interpret the visual guidance presented in the image. 

\noindent \textbf{Fine-tuning Results.} By fine-tuning models on our curated chart editing training set, all three models exhibit obviously performance gains. The fine-tuned Qwen2-vl shows remarkable gains in $\Delta$SSIM scores, increasing from 12.59 to \textbf{63.23} in the textual editing task and from 6.5 to \textbf{57.88} in the visual editing task, even surpassing GPT-4o. Similarly, the fine-tuned Llama-3.2-vision achieves the highest execution rate (\textbf{91.4\%}) on the textual setting and demonstrates exceptional Non-intervention Robustness (\textbf{83.4\%}). Fine-tuning also bridges the performance gap between the textual and visual settings, supplementing the key capabilities of the models to comprehend the visual editing instruction. This also demonstrates the effectiveness of our constructed training dataset.

\section{Analysis}

\subsection{Multi-level Task Analysis}
\begin{figure*}[h]
  \centering
  \includegraphics[width=\linewidth]{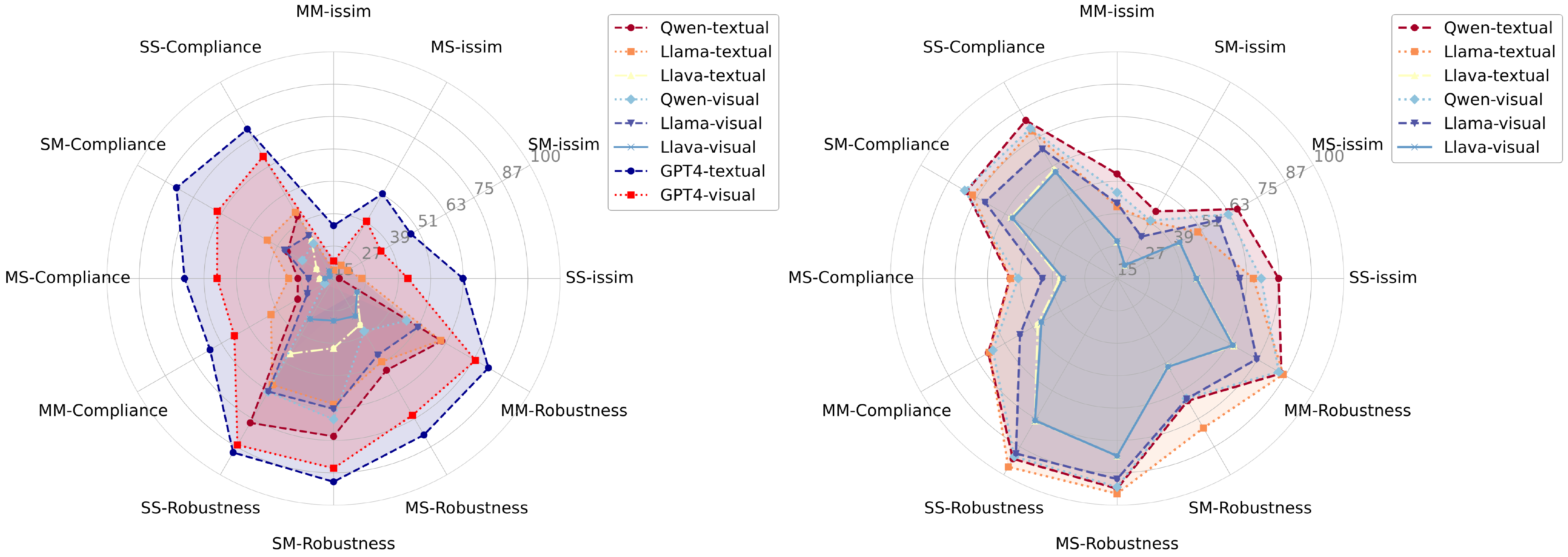}
  \caption{Performance of models on tasks with various difficulty levels. Left: Zero-shot results; Right: Results of models fine-tuned on our training set. The difficulty levels follow the order SM > MM > MS > SS.}
\end{figure*}


To comprehensively evaluate model performance, we structured our dataset into four categories based on modification targets and instructions. The data were split in a 3:3:3:1 ratio into the following groups: single-target single-instruction (SS), multi-target single-instruction (MS), single-target multi-instruction (SM), and multi-target multi-instruction (MM).
\begin{figure}[H]
  \centering
  \includegraphics[width=\linewidth]{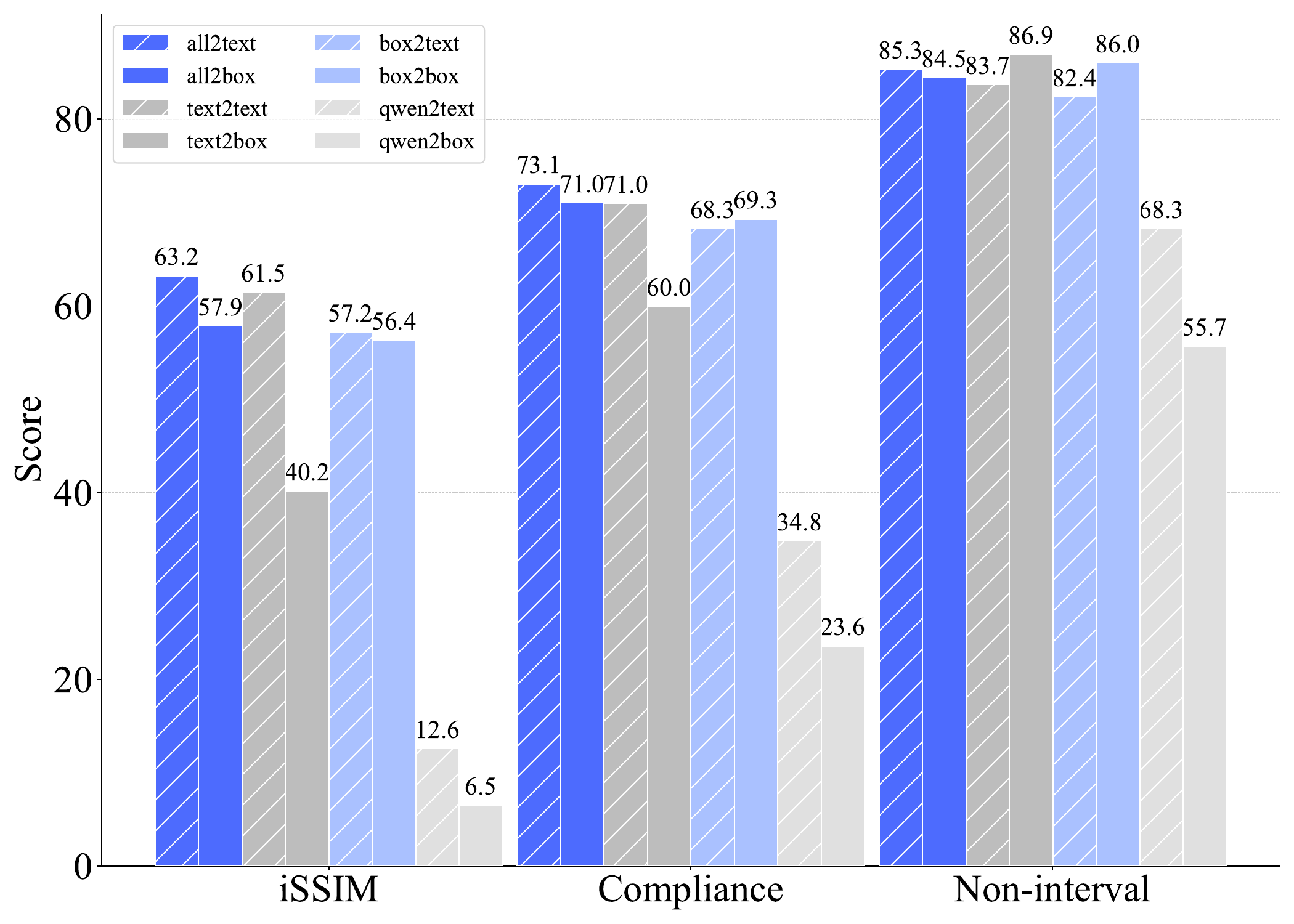}
  \caption{
  Model performance transfer results across textual-guided and visual (box) -guided settings. For example, \textit{all2text} denotes fine-tuning the model on both tasks and evaluating it on the textual-guided task setting.
  }
  \label{fig:ablation}
\end{figure}
We find that model performance declines as the number of modification targets and instructions increases. Notably, an increase in the number of instructions leads to a greater drop in performance compared to an increase in targets.
Before fine-tuning, models generally perform worst in the multi-target multi-instruction (MM) scenario. This suggests that it is difficult for models to track multiple visual elements and apply multiple modifications simultaneously. However, after fine-tuning, models perform worst in the single-target multi-instruction (SM) scenario. This may be due to the difficulty of applying multiple modifications simultaneously to a single target and get them all correct.


\subsection{Ablation Study}
\noindent To understand how different training components contribute and to investigate cross-task transfer effects, we conduct ablation studies using the Qwen2-vl model under three training configurations: training on both tasks (All), text-guided task only (Textual-only), and visual indicator-guided task only (Visual-only). Figure \ref{fig:ablation} summarizes the performance results for both evaluation tasks.

\noindent Models trained jointly on both tasks consistently outperform models trained on a single task, achieving the best scores on most metrics ($\Delta$SSIM: 63.23/57.88, Compliance: 73.06/71.04). This indicates beneficial knowledge transfer from joint training. Additionally, we observe asymmetric transfer effects. The Textual-only model achieves reasonable performance on text-guided tasks ($\Delta$SSIM: 61.46) but struggles significantly with visual indicator-guided tasks ($\Delta$SSIM: 40.17). In contrast, the Visual-only model shows more balanced performance across both task types ($\Delta$SSIM: 57.18/56.36). This asymmetry suggests that learning from visual instructions effectively transfers to text-based tasks, but the reverse is limited. These findings underline the importance of multi-task training for robust chart editing, especially when models must handle both visual and textual instructions in practical settings.

\begin{figure*}[h] 
  \centering
  \includegraphics[width=\textwidth]{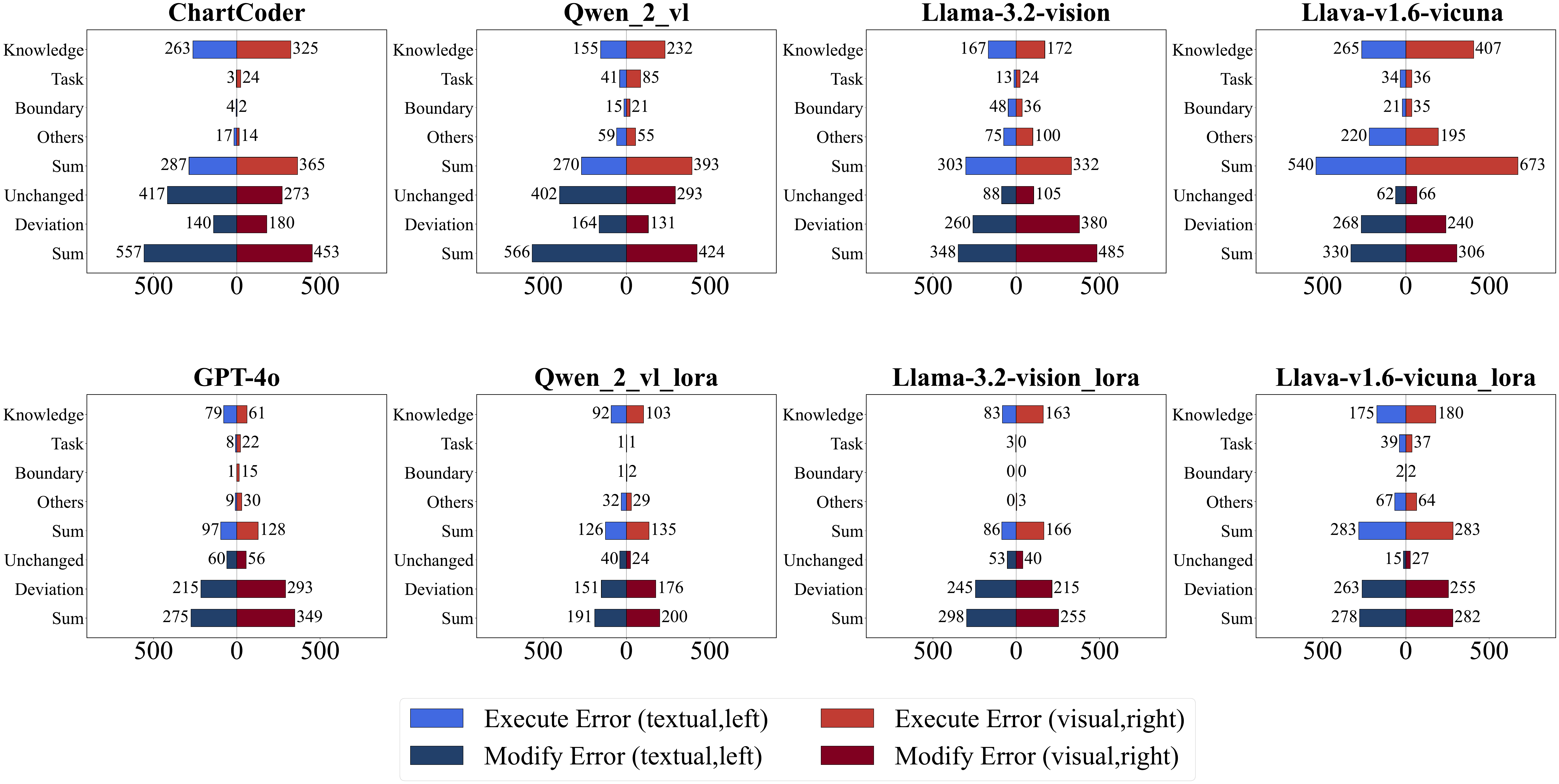} 
  \caption{Error analysis across models for both text description-guided (left) and bounding box-guided (right) tasks reveals key insights. Execution errors are categorized into: Knowledge (expert knowledge deficiency), Task (task comprehension failures), and Boundary (boundary violations). Modification errors include two types: Unchanged (failure to apply necessary edits) and Deviation (incorrect or excessive changes). Notably, fine-tuned models demonstrate substantial reductions in all error categories compared to their base counterparts, indicating improved task understanding and editing precision.} 
  \label{} 
\end{figure*}

\subsection{Error Analysis}

In our investigation, we perform a detailed error analysis across eight multimodal models. Our goal was to identify the key limitations of existing models and assess improvements brought by our specialized fine-tuned model. We categorized errors into two main groups:

\noindent \textbf{Execution Errors}: These occur when the generated code fails to execute, resulting in no chart output. Execution errors can be subdivided as follows:
\begin{enumerate}
\item \textbf{Expert Knowledge Deficiency}: Models lack essential technical knowledge about matplotlib, such as neglecting required library imports or using incorrect function calls.
\item \textbf{Task Comprehension Failures}: Models produce incomplete code loops or code exceeding token limits, reflecting a misunderstanding of task requirements.
\item \textbf{Boundary Violation Exceptions}: Models fail to properly align visual elements with corresponding code, leading to alignment errors.
\item \textbf{Miscellaneous Errors}: Errors that do not fit clearly into the above categories.
\end{enumerate}

\noindent \textbf{Modification Errors}: These happen when the generated code executes successfully but does not produce the desired visual changes ($\Delta$SSIM = 0). Modification errors include:
\begin{enumerate}
\item \textbf{Unchanged Failures}: Models generate executable code that leaves the original chart unchanged, often due to poor alignment between instructions and relevant code segments.
\item \textbf{Deviation Errors}: Modifications occur but differ from intended results, increasing the visual difference from the ground truth. These errors typically arise from misunderstandings of modification intent or incorrect parameter adjustments.
\end{enumerate}

Our evaluation of vision-language models on the ChartM$^3$ benchmark reveals key challenges and opportunities for improving chart editing capabilities. We observe substantial performance differences among models: proprietary models like GPT-4o performed notably better, whereas many open-source models~\cite{liu2024llavanext,Lee2025TrimmedLlama,wang2024qwen2vl,Zhao2025ChartCoder} struggled with both execution accuracy and semantic interpretation. Our error analysis further identifies semantic interpretation as the primary challenge, particularly in visual indicator-guided tasks. The persistent performance gap between textual and visual tasks highlights a core limitation of current models to interpret visual cues in structured visualizations.

Nevertheless, our fine-tuning approach effectively addresses these limitations. The fine-tuned models significantly reduce both execution and modification errors, surpassing GPT-4o on certain metrics. These results highlight that targeted training can greatly improve models’ ability to translate visual intent into accurate code implementations for data visualization tasks.


\subsection{Correlation with Human Evaluation}

To validate the effectiveness of our proposed automatic evaluation metrics, we conduct a human preference assessment to measure the alignment between human judgments and our metrics. We collect 200 samples, each evaluated by three independent assessors. Evaluators were required to compare outputs from various models using the provided modification instructions, targets, original images, and ground truth references. They then selected the better result based on three aspects: (1) Which model’s prediction appears visually closer to the ground truth? (2) Which model better edits the specified regions according to the editing instruction? (3) Which model better preserves the regions that are not supposed to be edited? We determined the final results for each sample by majority vote (See Appendix for more details). The results are shown in Table \ref{tab:freq}, where our $\Delta$SSIM metric and Compliance score demonstrate high agreement with human judgement, indicating the reliability of our proposed evaluation framework.

\begin{table}
  \caption{Agreement between the automatic metrics and human evaluation.}
  \label{tab:freq}
  \begin{tabular}{cc}
    \toprule
    Automatic Metrics&Human Agreement\\
    \midrule
    $\Delta$SSIM & 0.86 \\
    Directive Compliance Ratio & 0.70\\
    Non-intervened Robustness & 0.56\\
  \bottomrule
\end{tabular}
\end{table}
\section{Conclusion}
In this research, we introduced the ChartM$^3$ benchmark, which evaluates multimodal large language models' capabilities through chart editing tasks. ChartM$^3$ focuses on modifications to specific objects within charts, assessing models' abilities in cross-modal understanding, reasoning, and code generation with fine-grained precision. We implemented two different editing approaches with varying difficulty levels, providing a more comprehensive and in-depth evaluation. Our benchmark contributes to the advancement of AGI by establishing new standards for multimodal interaction. Despite our novel and comprehensive design, ChartM$^3$ has certain limitations. In real-world scenarios, modification instructions are often parameter-agnostic, with users typically providing general guidance rather than parameter-level instructions—a context where intelligent parameter selection better aligns with AGI aspirations. Additionally, because only Matplotlib among mainstream charting libraries supports the flexible, object‐level edits our tasks require, and the limited chart-to-code capabilities of existing open-source models, we provided code based on matplotlib to ensure modification quality, which restricts our evaluation to Matplotlib. We believe that ChartM$^3$ advances the field of chart editing—a practical domain where general models currently underperform, thus accelerating progress toward artificial general intelligence. Future research could explore more ambitious directions, such as eliminating provided code or enabling direct image editing without code involvement.

\section*{Acknowledgements}
We thank all reviewers for their insightful comments and suggestions. 
This work was partially supported by the Beijing Natural Science Foundation (No. L233008).

\balance
\bibliographystyle{ACM-Reference-Format}
\bibliography{sample-base}



\end{document}